\documentclass[review]{elsarticle}

\usepackage[english]{babel}
\usepackage[T1]{fontenc}
\usepackage{etoolbox}
\usepackage{float}
\makeatletter

\def\ps@pprintTitle{%
   \let\@oddhead\@empty
   \let\@evenhead\@empty
   \let\@oddfoot\@empty
   \let\@evenfoot\@oddfoot
}
\makeatother
\usepackage{comment}

\usepackage{subfig}
\usepackage{hyperref}
\usepackage{amsmath}

\usepackage{amsmath,amssymb,amsfonts}
\usepackage{algorithmic}
\usepackage[]{graphicx}
\usepackage{lipsum}
\usepackage{balance}
\usepackage{textcomp}
\usepackage{xcolor}
\usepackage{multirow}
\usepackage{color}
\usepackage{lscape}

\begin{document}

\begin{frontmatter}

\title{Analysis of French Phonetic Idiosyncrasies \\for Accent Recognition} 

\author[add1]{Pierre Berjon }
\ead{pierre.berjon@etu.enseeiht.fr}
\author[add2]{Avishek~Nag}
\ead{avishek.nag@ucd.ie}
\author[add3,add4]{Soumyabrata~Dev\corref{mycorrespondingauthor}}
\cortext[mycorrespondingauthor]{Corresponding author. Tel.: + 353 1896 1797.}
\ead{soumyabrata.dev@ucd.ie}

\address[add1]{Department de Sciences du Numérique, INP-ENSEEIHT, Toulouse, France}
\address[add2]{School of Electrical and Electronic Engineering, University College Dublin, Ireland}
\address[add3]{ADAPT SFI Research Centre, Dublin, Ireland }
\address[add4]{School of Computer Science, University College Dublin, Ireland}

\begin{abstract}
Speech recognition systems have made tremendous progress since the last few decades. They have developed significantly in identifying the speech of the speaker. However, there is a scope of improvement in speech recognition systems in identifying the nuances and accents of a speaker. It is known that any specific natural language may
possess at least one accent. Despite the identical word phonemic composition, if it is pronounced in different accents, we will have sound waves, which are different from each other. Differences in pronunciation, in accent and intonation of speech in general, create one of the most common problems of speech recognition. If there are a lot of accents in language we should create the acoustic model for each separately. We carry out a systematic analysis of the problem in the accurate classification of accents. 
We use traditional machine learning techniques and convolutional neural networks, and show that the classical techniques are not sufficiently efficient to solve this problem. Using spectrograms of speech signals, we propose a multi-class classification framework for accent recognition. In this paper, we focus our attention on the French accent. We also identify its limitation by understanding the impact of French idiosyncrasies on its spectrograms.

\end{abstract}


\end{frontmatter}

\section{Introduction}

Accent recognition is one of the most important topics in automatic speaker and speaker-independent speech recognition (SI-ASR) systems in recent years. The growth of voice-controlled technologies has becoming part of our daily life, nevertheless variability in speech makes these spoken language technologies relatively difficult. One of the profound variability in a speech signal is the accent. Different models could be developed to handle SI-ASR by accurately classifying the various accent types~\cite{ref6}. Such a successful accent recognition module can be integrated into a natural language processor, leading to its wide ranging impact in finance~\cite{islam2020foreign}, medical science~\cite{raji2020decision}, and sustainable environment~\cite{wu2021ontology}. 

Dialect/accent refers to the different ways of pronouncing/speaking a language
within a community. Some illustrative examples could be American English versus British English speakers or the Spanish speakers in Spain versus Caribbean. During the past few years, there have been significant attempt to automatically recognize the dialect or accent of a speaker given his or her speech utterance. Recognition of dialects or accents of speakers prior to automatic speech recognition (ASR) helps in improving performance of
the ASR systems by adapting the ASR acoustic and/or language models appropriately. Moreover, in applications such as smart assistants
as the ones used in smartphones, by recognizing the accent of the caller and then connecting the caller to agent with similar dialect or accent will produce more user-friendly environment for the users of the application.

Most of the existing techniques do not possess good accuracy in identifying the various accents. One of the reasons we are having trouble to have a good accuracy in the accent recognition problem is the lack of knowledge we have of English syllabic structure. In order to approximate English phonology, we have to understand the native language similarities of articulation, intonation, and rhythm. In the past, the research has focused on phone inventories and sequences, acoustic realizations, and intonation patterns. Therefore, it is important to study the English syllable structure. The main problem behind word recognition is the understanding of the syllable. It usually consists of an obligatory vowel with optional initial and final consonants. One familiar way of subdividing a syllable is into \textit{onset} and \textit{rhyme}. All syllables in all languages phonetically at least consist of onset and rhyme. However, these categories alone do not indicate where the syllable is placed within the word. In order to capture foreign accents in English, we want to highlight those constituents of the syllable that are most likely to prove difficult for speakers of languages in which they are not contained~\cite{scope}.

In this paper, we focus on the specifications of the French language. We are interested in identifying the 
idiosyncrasies~\cite{ref9} of French people that lead a model into predicting the wrong accent. 

\subsection{Related work}
Berkling \textit{et al.} \cite{scope} discussed 
the tonal and non-tonal languages and their treatment in speech recognition systems. In Kardava \textit{et al.} \cite{kardava}, they have developed  an  approach to  solve the above mentioned  problems  and  create  more  effective,  improved speech  recognition  system  of  Georgian  language  and  of languages,  that  are  similar  to  Georgian  language. Katarina \textit{et al.} proposed \cite{katarina}, an automatic method of detection of the degree of foreign accent and the results are compared with accent labeling carried out by an expert phonetician. In \cite{jouvet}, they give a new approach for modelling allophones in a speech recognition system based on hidden Markov models.

In \cite{ref12}, they studied mutual influences between native and non-native vowel production during learning, \textit{i.e.}, before and after short-term visual articulatory feedback training with non-native sounds. To obtain a speaker’s pronunciation characteristics, \cite{ref10} gave a method based on an idea from bionics, which uses spectrogram statistics to achieve a characteristic spectrogram to give a stable representation of the speaker’s pronunciation from a linear superposition of short-time spectrograms. Hossari \textit{et al.} in \cite{hossari2019test} used a two-stage cascading model using Facebook’s fastTex implementation~\cite{joulin2016bag} to learn the word embeddings. Davies \textit{et al.} presented advanced computer vision methods, emphasizing machine and deep learning techniques that have emerged during the past 5–10 years~\cite{ref8}. The book provides clear explanations of principles and algorithms supported with applications. In \cite{ref11}, Farris present the Gini index and several measures of integrity.

\subsection{Contributions of the paper}
The main contributions of this paper\footnote{With the spirit of reproducible research, the code to reproduce the results in this paper is shared at \texttt{\url{https://github.com/pberjon/Article-Accent-Recognition}}.} can be summarized as follows:

\begin{itemize}
    \item Highlighting the problem of the limit in the context of the study of accent recognition. In this paper, we will show there exists a ``natural'' limit of the accuracy when it comes to accent classification. The main aim of this work will be to address that limit and give a solution to that problem.
    \item Highlighting French idiosyncrasies restricting the accuracy values of deep learning models. In this paper, we focused our work on the French speakers. We decided to study the language habits of French speakers that could explain the decrease in precision. Indeed, the English language is an Indo-European Germanic language while the French is a Latin language, which means that their structure is very different. Thus, we will find strongly similar words between the two languages, but the way of pronouncing them will often vary a lot. Thus, the study of these Latin habits is particularly interesting in the context of our work: understanding which aspects of the French language reduce the effectiveness of our models will allow us to better recognize a French accent later on.
    \item Highlighting the incidence of these idiosyncrasies in the spectrograms, and therefore the models in question. Once we have isolated more clearly the responsible French idiosyncrasies, we determine their real impact on the models used (CNN in our case) by the precise study of spectrograms of vocal samples used. In this case, we will compare different spectrograms for the same sentence and determine the differences between a ``French'' and ``English'' spectrogram, for a specific idiosyncrasy. 
\end{itemize}

The rest of the paper is structured as follows. Section 2 discusses the data and the methods we used in our preliminary study (dataset and neural networks) and Section 3 discusses results we obtained with these methods. In Sections 4 and 5, we analysed the French speakers idiosyncrasies and their consequences on spectrograms. Finally, Section 6 concludes the work and discusses our future works.

\section{A primer on French speakers idiosyncrasies}
\label{sec:headings}
In this section, we provide a primer to the readers on the various types of speech idiosyncrasies exhibited by French speakers. 

\subsection{French-infused vowels}

Nearly every English vowel is affected by the French accent \cite{ref12}. French has no diphthongs, so vowels are always shorter than their English counterparts. The long A, O, and U sounds in English, as in say, so, and Sue, are pronounced by French speakers like their similar but un-diphthonged French equivalents, as in the French words sais, seau, and sou. For example, English speakers pronounce say as [seI], with a diphthong made up of a long "a" sound followed by a sort of "y" sound. But French speakers will say [se] - no diphthong, no "y" sound. English vowel sounds which do not have close French equivalents are systematically replaced by other sounds, as it's showed in Table \ref{tab:4}.

\begin{table}[htb]
\centering
\begin{tabular}{|*{10}{c|}}
    \hline
    \textbf{Usual English pronunciations and French pronunciation} \tabularnewline
    \hline
    short A, as in fat\\
    \texttt{French Accent :} pronounced "ah" as in father\tabularnewline
    \hline
    long A followed by a consonant, as in gate\\
    \texttt{French Accent :} pronounced like the short e in get\tabularnewline
    \hline
    ER at the end of a word, as in water\\
    \texttt{French Accent :} pronounced air \tabularnewline
    \hline
    short I, as in sip\\
    \texttt{French Accent :} pronounced "ee" as in seep \tabularnewline
    \hline
    long I, as in kite\\
    \texttt{French Accent :} elongated and almost turned into two syllables: [ka it] \tabularnewline
    \hline
    short O, as in cot\\
    \texttt{French Accent :} pronounced either "uh" as in cut, or "oh" as in coat \tabularnewline
    \hline
    U in words like full\\
    \texttt{French Accent :} pronounced "oo" as in fool \tabularnewline
    \hline
\end{tabular}
\caption{\label{tab:4}Highlighting of French main mispronunciations of the English language.}
\end{table}
 
\subsection{Dropped Vowels, Syllabification, and Word Stress
}
French people pronounce all schwas (unstressed vowels). Native English speakers tend toward "r'mind'r," but French speakers say "ree-ma-een-dair." They will pronounce amazes "ah-may-zez," with the final e fully stressed, unlike native speakers who will gloss over it: "amaz's." And the French often emphasize the -ed at the end of a verb, even if that means adding a syllable: amazed becomes "ah-may-zed."

Short words that native English speakers tend to skim over or swallow will always be carefully pronounced by French speakers. The latter will say "peanoot boo-tair and jelly," whereas native English speakers opt for pean't butt'r 'n' jelly.

Because French has no word stress (all syllables are pronounced with the same emphasis), French speakers have a hard time with stressed syllables in English, and will usually pronounce everything at the same stress, like actually, which becomes "ahk chew ah lee." Or they might stress the last syllable - particularly in words with more than two: computer is often said "com-pu-TAIR."

\subsection{French-accented Consonants
}
H is always silent in French, so the French will pronounce happy as "appy.". Once in a while, they might make a particular effort, usually resulting in an overly forceful H sound - even with words like hour and honest, in which the H is silent in English. J is likely to be pronounced "zh" like the G in massage. R will be pronounced either as in French or as a tricky sound somewhere between W and L. Interestingly, if a word starting with a vowel has an R in the middle, some French speakers will mistakenly add an (overly forceful) English H in front of it. For example, arm might be pronounced "hahrm."

TH's pronunciation will vary, depending on how it's supposed to be pronounced in English:
\begin{itemize}
    \item voiced TH [ð] is pronounced Z or DZ: "this" becomes "zees" or "dzees"
    \item unvoiced TH is pronounced S or T: "thin" turns into "seen" or "teen"
\end{itemize}

Letters that should be silent at the beginning and end of words (psychology, lamb) are often pronounced.


\section{Accent Recognition System}

\subsection{Features for detecting accents}
Spectrograms are pictorial representation of sound we can use for speech recognition \cite{ref10}. The x-axis represents time in seconds while the y-axis represents frequency in Hertz. Different colors represent the different magnitude of frequency at a particular time. We can think of the spectrogram as an image. Once the audio file is converted to an image, the problem reduces to an image classification task. Based on the number of images, algorithms like Support Vector Machines (SVM), \textit{etc.} are used to classify sound, validate the speaker.

\begin{figure}[htb]
\centering
\subfloat[\centering Signal]{{\includegraphics[width=0.8\textwidth]{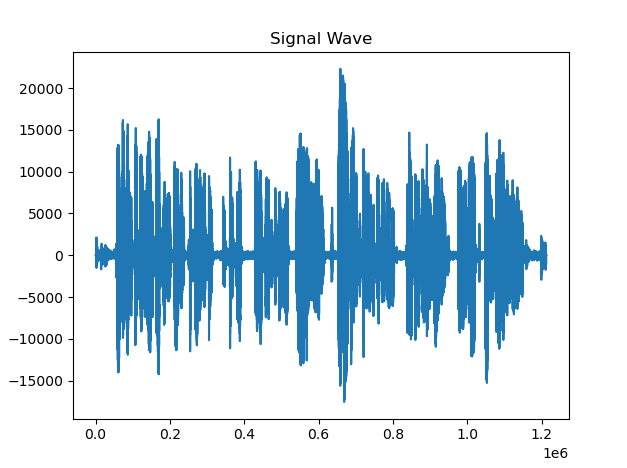} }}\\
\subfloat[\centering Spectrogram]{{\includegraphics[width=0.9\textwidth]{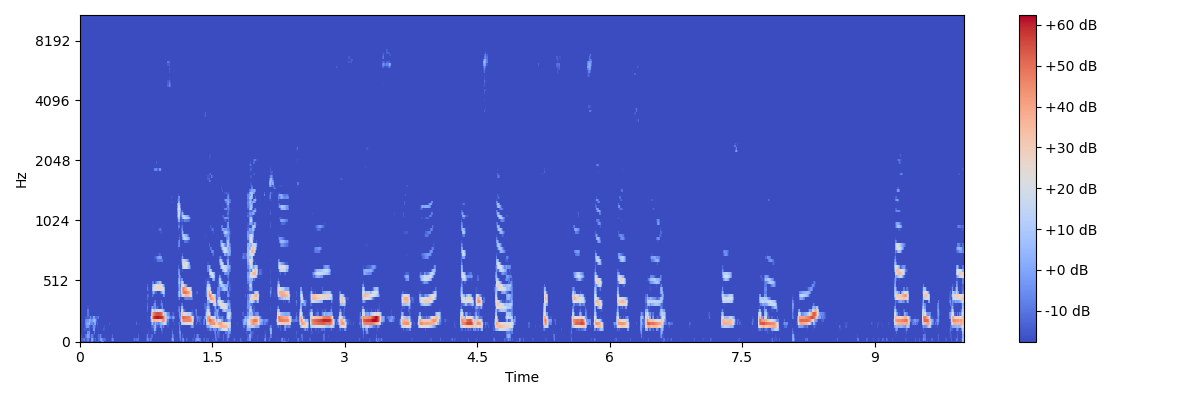} }}%
\caption{Signal and spectrogram of a french accent sample}%
\label{fig:example}%
\end{figure}

\subsection{Our proposed framework for detecting accents}
We used different Machine Learning and Deep Learning models, and the first one is a two convolutional layers neural network with 5 different accents as shown in Figure 2. This neural network is a 2-layer Convolutional Neural Network : one with 32 filters and a ReLu activation function, and another one with 64 filters and a ReLu activation function.


\begin{figure}[htb]
\centering
\includegraphics[width=0.95\textwidth]{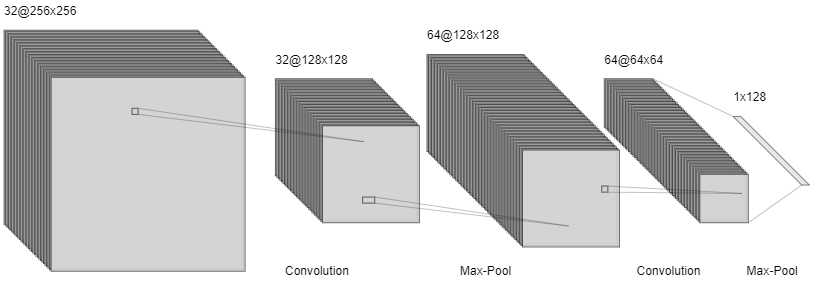}
\caption{CNN with 2 layers with ReLu activation function}
\label{fig:2cnn}
\end{figure}

We will focus on this 2-layer CNN for the rest of our work.

\section{Results and Discussion}

\subsection{Dataset}

Everyone who speaks a language, speaks it with an accent. A particular accent essentially reflects a person's linguistic background. When people listen to someone speak with a different accent from their own, they notice the difference, and they may even make certain biased social judgments about the speaker. In this paper, we used the Speech Accent Archive \cite{weinberger}. It has been established to uniformly exhibit a large set of speech accents from a variety of language backgrounds. Native and non-native speakers of English all read the same English paragraph and are carefully recorded.

\begin{figure}[htb]
\centering
\includegraphics[width=0.9\textwidth]{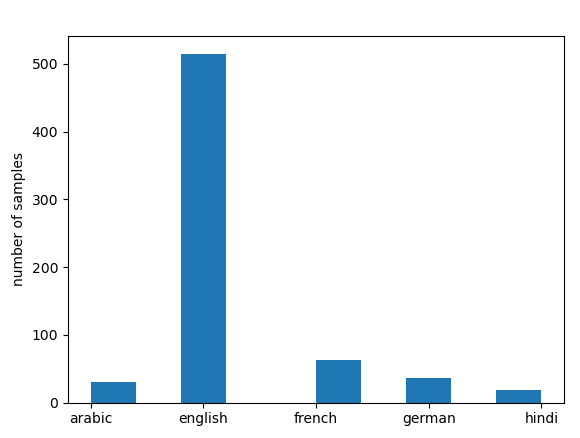}
\caption{The distribution of the samples across the five languages in the dataset.}
\label{fig:distribution}
\end{figure}


This dataset allows us to compare the demographic and linguistic backgrounds of the speakers in order to determine which variables are key predictors of each accent. The speech accent archive demonstrates that accents are systematic rather than merely mistaken speech. It contains 2140 speech samples, each from a different talker reading the same reading passage. Talkers come from 177 countries and have 214 different native languages. Each talker is speaking in English. The samples were collected by many individuals under the supervision of Steven H. Weinberger, the most up-to-date version of the archive is hosted by George Mason University and can be found here : \texttt{\url{https://www.kaggle.com/rtatman/speech-accent-archive}}. \cite{weinberger}

\subsection{Accent recognition metric}
In order to provide an objective evaluation of the accent recognition task, we compute the overall accuracy, F1-macro, F1-micro and hamming loss~\cite{dev2019identifying}. These metrics are defined as:


\begin{equation*}
\begin{split}
    ACC & = \frac{tp + tn}{(tp + fp) + (tn + fn)} \\
    F1_{macro} & = \frac{1}{N}\sum_{i=1}^{N} F1_{i} \\
    F1_{micro} & = 2 \frac{\textrm{}{Micro-precision * Micro-recall}}{\textrm{Micro-precision + Micro-recall}}\\
    HL & = \frac{1}{NL} \sum_{i=1}^{N} \sum_{l=1}^{L} Y_{i,l} \oplus X_{i,l}
\end{split}
\end{equation*}

In the overall accuracy formula, tp, tn, fp, fn stand respectively for true positive, true negative, false positive and false negative. In the Hamming loss formula, 
$\oplus$ denotes exlusive-or, $X_{i,l}$ ($Y_{i,l}$) stands for boolean that the $i^{\textrm{th}}$ datum ($i^{\textrm{th}}$prediction) contains the $l^{\textrm{th}}$ label


Table~\ref{tab:accuracy-table} demonstrates the evaluation metric obtained via SVM technique and two variants of CNN model. 

\begin{table}[H]
\centering
\begin{tabular}{|*{5}{c|}}
    \hline
    \multicolumn{5}{|c|}{\textbf{Comparison of SVM and CNNs}} \\
    \hline
    \textbf{Model} & \textbf{Overall ACC} & \textbf{F1 Macro} & \textbf{F1 Micro} & \textbf{Hamming Loss} \tabularnewline
    \hline
    SVM & 0.3518 & 0.33458 & 0.33458 & 0.38043 \tabularnewline
    \hline
    2-layer CNN & 0.70652 & 0.405 & 0.70652 & 0.29348 \tabularnewline
    \hline
    4-layer CNN & 0.6529 & 0.52 & 0.73913 & 0.26087 \tabularnewline
    \hline
\end{tabular}
\caption{Average accent classification accuracy across the different languages using various benchmarking models.}
\label{tab:accuracy-table}
\end{table}

With regular machine learning methods as SVM, we obtained low accuracy of 0.35. As expected, the impact of Deep Learning methods \cite{ref8} is quite clear here. We observe from Table~\ref{tab:accuracy-table}, that the Convolutional Neural Networks achieves an accuracy of 
$0.65$. However, we observe that we do not obtain an optimal score if we use too many layers in our model. Depending upon how large our dataset is, the CNN architecture is implemented. Adding layers unnecessarily to any CNN will increase our number of parameters only for the smaller dataset. It’s true for some reasons that on adding more hidden layers, it will give a better accuracy. That's true for larger datasets, as more layers with less stride factor will extract more features for the input data. In CNN, how we play with the architecture is completely dependent on what our requirement is and how our data is. Increasing unnecessary parameters will only overfit your network, and that's the reason why our CNN with 2 layers has better results than with 4.

A macro-average will compute the metric independently for each class and then take the average (hence treating all classes equally), whereas a micro-average will aggregate the contributions of all classes to compute the average metric. In a multi-class classification setup, micro-average is preferable if we suspect there might be class imbalance issue (\textit{i.e.} we may have many more examples of one class, as compared to other classes). Table 2 explains this scenario clearly. We observe that neural networks show better F1-score values in the context of multi-class classification. In such situation, Hamming Loss is a good measure of model performance. The lower the Hamming loss, the better is the model performance. In our case, Hamming loss ranges from 0.26 till 0.39, which is considered as good results, especially in the context of 5-class multi-class classification problem. 

\subsection{Multi-class accent recognition metric}

In this case of multi-class classification, we are considering ACC, AGF, AUC and GI.  

\begin{equation*} \label{eq1}
\begin{split}
ACC & = \frac{tp + tn}{tp + tn + tp + tn} \\
AGF & = (1 + \beta^2) \frac{\textrm{precision} * \textrm{recall}}{(\beta^2 * \textrm{precision}) + recall}\\
AUC & = \frac{\textrm{recall} + \textrm{sensibility}}{2}\\
GI & = 1 - \sum_{j=1}^{n} p_j^2 = 1
\end{split}
\end{equation*}

We obtained these results in the confusion matrices with the 2-layer CNN and the SVM method:

\begin{table}[htb]
\centering
\begin{tabular}{|*{5}{c|}}
    \hline
    \multicolumn{5}{|c|}{\textbf{SVM}} \\
    \hline
    \textbf{Classes} & \textbf{ACC} & \textbf{AGF} & \textbf{AUC} & \textbf{GI} \tabularnewline
    \hline
    English & 0.42391 & 0.21774 & 0.36781 & -0.26437 \tabularnewline
    \hline
    Arabic & 0.71739 & 0.0 & 0.5 & 0.0 \tabularnewline
    \hline
    French & 0.34783 & 0.51315 & 0.49171 & -0.01658 \tabularnewline
    \hline
    German & 0.92391 & 0.0 & 0.5 & 0.0 \tabularnewline
    \hline
    Hindi & 0.95652 & 0.0 & 0.5 & -0.01124 \tabularnewline
    \hline
\end{tabular}
\caption{Multi-class classification metric values using the SVM model.}
\label{tab:2}
\end{table}

Table~\ref{tab:2} indicates that the 
results for Arabic, Hindi and German accents are better. This can easily be explained by the size of the data sets corresponding to each accent. 
This is a result that shows fairly well the limit of \textit{classical} machine learning algorithms. This limit in the evaluation scores for classical machine learning models are also observed in the broad areas of network security~\cite{elsayed2019machine} and computer vision~\cite{jain2021using}. In this specific application of accent recognition, we observe that an increase in the number of vocal samples do not lead to an increased accuracy values. 
This difference is due to the lack of capacity of the SVM which has difficulty processing information as complex as images. 

\begin{table}[H]
\centering
\begin{tabular}{|*{5}{c|}}
    \hline
    \multicolumn{5}{|c|}{\textbf{2-layer CNN}} \\
    \hline
    \textbf{Classes} & \textbf{ACC} & \textbf{AGF} & \textbf{AUC} & \textbf{GI} \tabularnewline
    \hline
    English & 1.0 & 1.0 & 1.0 & 1.0 \tabularnewline
    \hline
    Arabic & 0.95 & 0.71 & 0.74 & 0.48  \tabularnewline
    \hline
    French & 0.85 & 0.84 & 0.84 & 0.69  \tabularnewline
    \hline
    German & 0.84 & 0.80 & 0.80 & 0.61  \tabularnewline
    \hline
    Hindi & 0.87 & 0.32 & 0.53 & 0.06  \tabularnewline
    \hline
\end{tabular}
\caption{\label{tab:3} Multi-class classification metric values using our proposed 2-layer CNN model.}
\end{table}

Table~\ref{tab:3} indicates that 
the results are much more harmonized between the different accents. We still do not have a perfect match between the size of the dataset and the performance of the model, but the disparities between accents disappear.

We can observe from Table~\ref{tab:2} and Table~\ref{tab:3} that the \textit{classical} machine learning methods are quite ineffective and that the deep learning methods stand out clearly in accent recognition; that's why we will use the 2-layer CNNs as a reference for the rest of the paper. In most case, the SVM method is not powerful enough for us to have a good accuracy. That can be explained with the results we obtained on the Gini Index~\cite{ref11}. The values obtained by the index are quite low (negative values are considered quite low positive values), which means that in the case of SVM, the 
spectrograms are similar in nature. Such SVM methods are not selective enough to clearly determine the accent (which is also shown by the AGF values). However, the SVM method is not totally to be excluded: in the context of the Hindi accent or the German accent, the SVM turns out to be more effective than all the deep learning methods used.

The total computing time is 1 minute and 23 seconds when our proposed model is executed on Google Colab using GPU. 

\section{Impact of Idiosyncrasies on Speech Spectrograms}

We will now study the idiosyncrasies of the French language and how it impacts the corresponding spectograms of the speech signals.

The spectrogram is a representation allowing to observe the whole of the decomposition spectral voice and speech on the same graphic representation. This tool is precise, informative and reliable to analyze the characteristics of sound production. In a first-cut analysis, we associate the spectrogram with the temporal pace, the power profile
and segmentation. More extensively, there are a significant number of
indicators, metrics and tools. This includes the fundamental frequency and its
derivatives, the alteration of voice and speech, and more generally the assessment of intelligibility. It is its ability to measure vocal alteration that will interest us here. 
We will focus on primarily two pieces of information given by the spectrogram: amplitude and frequency in our study.

\subsubsection{The un-diphthonged “y”}

Firstly, we will analyze differences on the spectrograms for the word “Wednesday”, where the French speaker is not supposed to use the “y” sound, like it was explained in French-infused Vowels. Here are the spectrograms of  an English speaker and a French speaker of the sentence “and we will go meet her Wednesday at the train station” :

\begin{figure}[htb]
\centering
\includegraphics[width=0.95\textwidth]{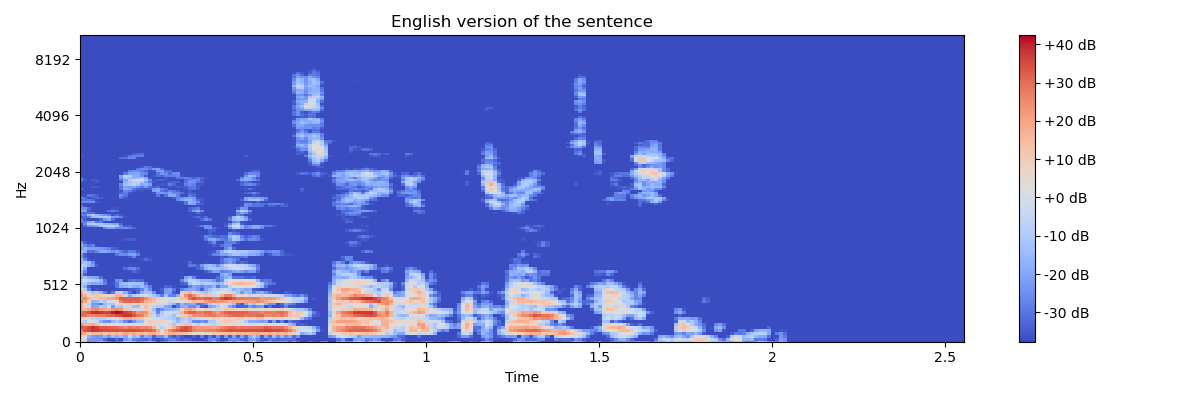}
\caption{“Wednesday” in English version : 0.8s-1.4s.}
\end{figure}

\begin{figure}[htb]
\centering
\includegraphics[width=0.95\textwidth]{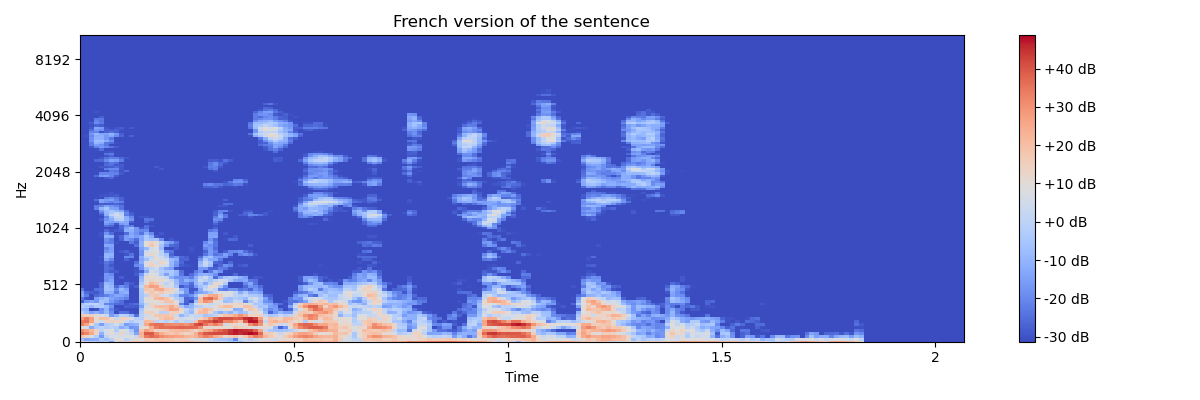}
\caption{“Wednesday” in French version : 0.7s-1.10s.}
\end{figure}

We can see, as expected, that at the end of the word (1.3-1.4 for English and 1.05-1.1 for French), the “y” is almost not even pronounced by the French speaker, while the English speaker pronounced it clearly. Indeed, the frequencies used are relatively similar on the whole of the audio sample, but certain syllables are \textit{pressed} with a much higher frequency by a French speaker. Consequently, the corresponding amplitude will be low in magnitude. This explains a clear difference between the perception of a word between a French speaker and an English speaker: the non-native will tend to pronounce English less loudly, but will support certain syllables much more than an English speaker.

\subsubsection{Voiced TH [ð] is pronounced Z or DZ}

French people tend to say "zees" instead of “these”. That’s what we can see in the sentence “Please call Stella, ask her to bring these things from the store.”.

\begin{figure}[htb]
\centering
\includegraphics[width=0.95\textwidth]{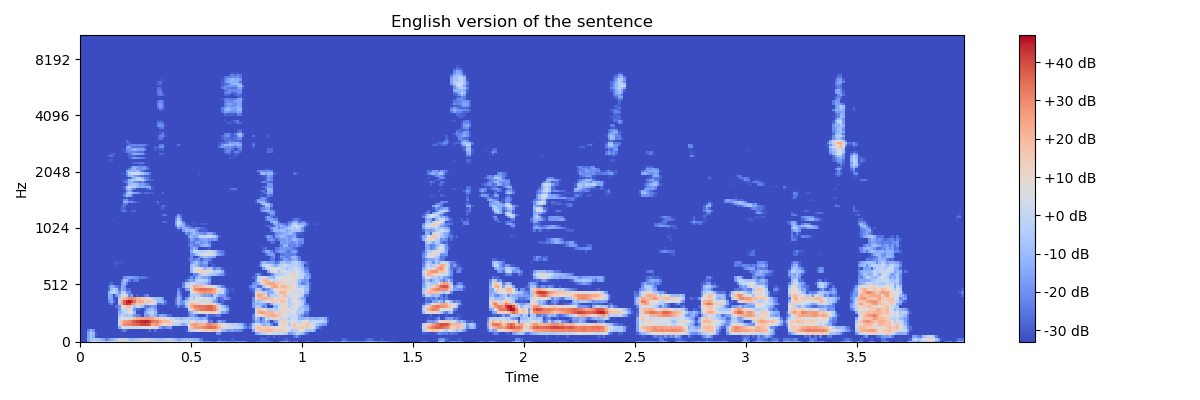}
\caption{“Bring these” in English version : 2.5s-3s.}
\end{figure}

\begin{figure}[htb]
\centering
\includegraphics[width=0.95\textwidth]{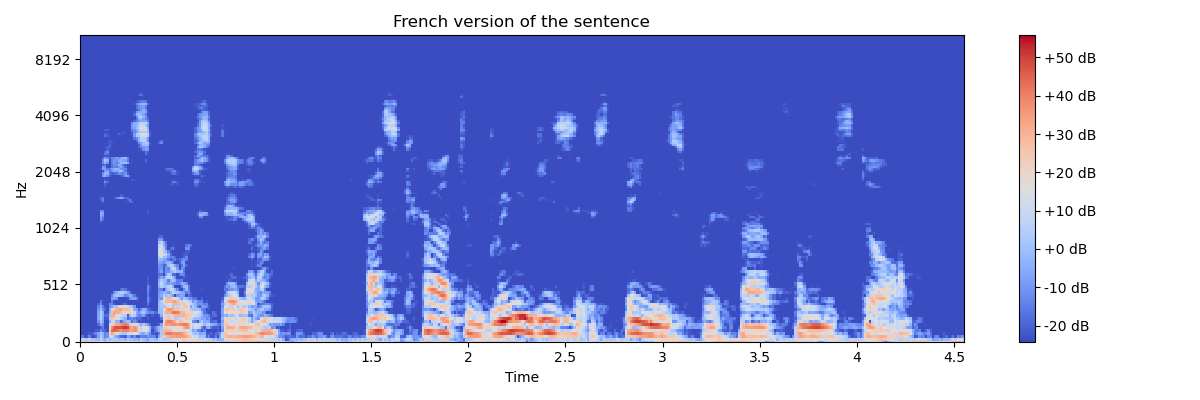}
\caption{“Bring these” in French version : 2.6s-3.2s.}
\end{figure}

It’s quite complicated to delimit the word “these” in this sentence because it is quite quick, so we will delimit “bring these”, as the word “bring” does not represent a major problem for French speakers. 

Here, we see that french people tend to diminish the importance of the word “bring” but accentuate the word “these”,  whereas English speakers seem to pronounce the sequence “bring these” at the same frequency. we think that’s why, for French speakers, the “th” sounds like “z”. Indeed, the closest sound to “th” is “z” in the French language, so it’s only natural for us to use it. Nevertheless, we think the reason why they accentuate it (because we could just use the sound “z” more discreetly) is because of the role of words like “these”, “the”, “this”... They’re articles, and in the French language, they tend to accentuate the most important parts of the sentence, which made this French speaker diminish “bring”, and accentuate “these”.

Thus, French speakers idiosyncrasies have a direct impact on audio samples spectrograms. Then, we can easily understand why these idiosyncrasies have a direct impact on the results of deep learning models: the first reason why we use spectrograms in order to develop Speech Recognition Systems is to turn an audio classification problem into an image classification problem. Then, if the idiosyncrasies of a specific language have that much effect on spectrograms, that means that the different languages have different spectrograms and this should help the deep learning models to get a better classification between English and French.

\section{Conclusions and future work}
In this paper, we have concluded that the classical deep learning models are not powerful enough to accurately predict the accent of an user. Therefore, we decided to study the differences between tonal and non-tonal languages, in order to clearly identify the obstacles that prevent us from achieving better results in accent recognition. To fulfill this purpose, we decided to devote our analysis on the French accent, which is a non-tonal language. In this paper, we studied the idiosyncrasies of French speakers: the characteristics of the spoken French language that have a direct impact on the pronunciation of English words by French speakers. In addition, we determined the consequences these idiosyncrasies have on spectrograms, and consequently on the accuracy of deep learning models. In the future, we would like to work further on the subject of French idiosyncrasies, by building a model which determines if an idiosyncrasy is present in an audio sample or not. This would allow us to more easily determine the presence of a French accent in an audio sample. Such accurate recognition of accents in a speech signal will lead to better automatic speech recognition systems. 

\section*{Acknowledgments}
The ADAPT Centre for Digital Content Technology is funded under the SFI Research Centres Programme (Grant 13/RC/2106\_P2) and is co-funded under the European Regional Development Fund.

\end{document}